\documentclass[lettersize,journal]{IEEEtran}
\usepackage{amsmath,amsfonts}
\usepackage{algorithmic}
\usepackage{algorithm}
\usepackage{array}
\usepackage[caption=false,font=normalsize,labelfont=sf,textfont=sf]{subfig}
\usepackage{textcomp}
\usepackage{stfloats}
\usepackage{url}
\usepackage{verbatim}
\usepackage{graphicx}
\usepackage{cite}
\usepackage{booktabs}
\usepackage{multirow}
\usepackage{siunitx}
\usepackage{array}
\begin{document}

\title{Implicit Fuzzification via Bounded Noise Injection for Robust Medical Image Segmentation}

\author{Bisheng Tang, Zhangfeng Ma, ~\IEEEmembership{Member,~IEEE,}
Chuchu Zhai, Feng Dong, Yaoqun Wu, Ammar Oad, Yifei Peng
\thanks{Bisheng Tang is with the School of Information Science and Engineering, Provincial Key Laboratory of Informational Service for Rural Area of Southwestern Hunan, Shaoyang University, Shaoyang 422000, China(e-mail: ucas459@gmail.com). Chuchu Zhai is with the Xinshao County People's Hospital, Shaoyang 422900, China(e-mail:tsinghua1129@outlook.com). Zhangfeng Ma, Feng Dong, Yaoqun Wu, Oad Ammar and Yifei Peng are with with the School of Information Science and Engineering, Shaoyang University, Shaoyang 422000, China(emails:zhangfeng.ma@vip.126.com, dongfeng@hnsyu.edu.cn, wuyaoqun@hnsyu.edu.cn, ammar.oad@hnsyu.edu.cn, yfpeng6909@126.com). Chuchu Zhai is the corresponding author.}
\thanks{Manuscript received April 19, 2021; revised August 16, 2021.}}

\markboth{Journal of \LaTeX\ Class Files,~Vol.~14, No.~8, August~2021}%
{Shell \MakeLowercase{\textit{et al.}}: A Sample Article Using IEEEtran.cls for IEEE Journals}

\IEEEpubid{0000--0000/00\$00.00~\copyright~2021 IEEE}

\maketitle

\begin{abstract}
Image segmentation remains fundamentally limited by boundary ambiguity arising from sampling-induced information loss and inherent uncertainty in pixel-wise labeling. Although encoder--decoder architectures such as U-Net achieve strong performance, they often produce overconfident predictions that fail to capture transition-region ambiguity. To address this issue, we propose \textbf{NoiseUNet}, a simple yet effective framework that injects bounded perturbations into skip connections to regularize cross-scale feature fusion. This mechanism enforces robustness to local feature variations and promotes boundary-aware representations. Theoretically, the perturbation induces an implicit fuzzification effect, yielding soft, data-driven memberships without requiring explicit fuzzy modeling. We further introduce \textbf{ThyR}, a real-world thyroid ultrasound dataset with inherently ambiguous boundaries. Experiments demonstrate that NoiseUNet consistently improves both segmentation accuracy and boundary fidelity.
\end{abstract}

 
\section{Introduction}
\IEEEPARstart{I}{mage} segmentation is a fundamental problem in computer vision, aiming to assign a semantic label to each pixel and transform raw visual signals into structured and interpretable representations. It plays a critical role in applications such as medical image analysis \cite{ronneberger2015u,li2025u}, remote sensing \cite{li2024review,wu2025farmseg_vlm}, and industrial inspection \cite{zhang2025small,wang2025overview}. In recent years, encoder--decoder convolutional neural networks have become the dominant paradigm due to their strong representation capacity and end-to-end optimization capability.

\begin{figure}
    \centering
    \includegraphics[width=1\linewidth]{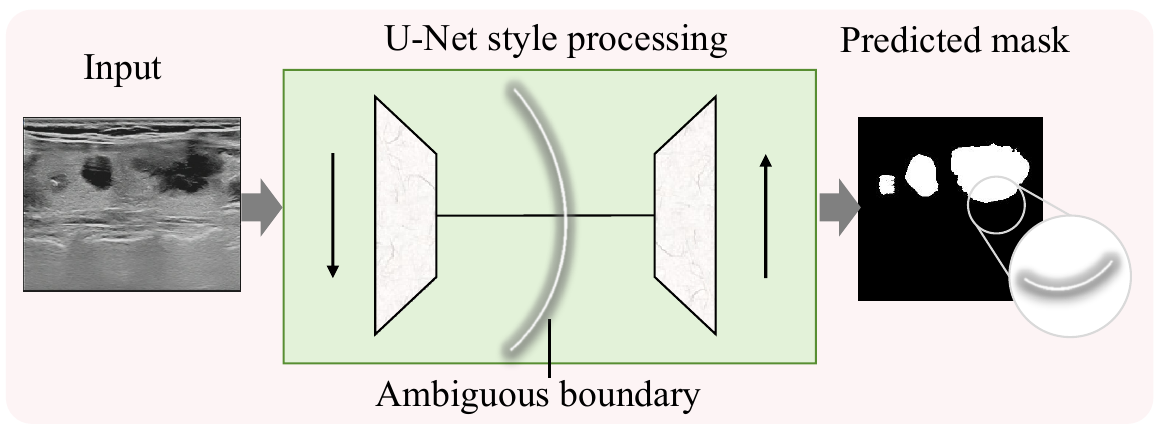}
    \caption{Boundary ambiguity in encoder--decoder segmentation. Down-sampling suppresses high-frequency boundary cues, and skip connections cannot fully recover the lost information, resulting in over-smoothed and uncertain predictions near object contours.}
    \label{fig:noise}
\end{figure}

Among these architectures, U-Net and its variants achieve remarkable success by combining hierarchical feature extraction with skip connections. The encoder progressively captures high-level semantic context through down-sampling, while the decoder restores spatial resolution via up-sampling and multi-scale feature fusion. Despite these advantages, accurate boundary delineation remains challenging, particularly in medical images with low contrast, noise, and blur. As illustrated in Fig.~\ref{fig:noise}, such degradations are closely related to the loss and distortion of boundary information during feature transformation.

A key cause of boundary degradation is sampling-induced ambiguity. Pooling and strided convolutions attenuate high-frequency components that encode structural discontinuities, which are essential for precise boundary localization. As a result, boundary features are disproportionately weakened during encoding. Although skip connections reintroduce high-resolution features, the information lost during down-sampling cannot be fully recovered, and the decoder tends to generate overly smooth interpolations, leading to blurred or imprecise segmentation outputs.

More fundamentally, boundary ambiguity reflects intrinsic uncertainty in pixel-to-label mapping rather than purely architectural limitations. In real imaging systems, boundaries often correspond to transition regions influenced by noise, blur, and discretization effects, where class membership is inherently ambiguous rather than strictly binary. However, most segmentation models rely on deterministic feature representations and crisp decision boundaries, which limits their ability to capture such uncertainty and results in sensitivity to small perturbations near object contours.

Noise injection offers a principled way to model this uncertainty within deep networks \cite{liao2024double, wang2025prescribed}. Instead of treating noise as an undesirable disturbance, it can act as a regularizer that encourages invariant and stable feature representations. In this work, we propose a noise-injected U-Net framework that introduces bounded perturbations along skip connections. By perturbing high-resolution features during cross-scale fusion, the model is encouraged to learn boundary-aware representations that remain stable under local variations, thereby reducing reliance on fragile high-frequency cues and improving robustness.

From a fuzzy systems perspective, this mechanism induces an implicit fuzzification process. Injecting bounded perturbations expands each deterministic feature into a local neighborhood in latent space, leading to soft, data-driven class memberships in boundary regions. Unlike classical fuzzy segmentation approaches that rely on predefined membership functions or explicit inference rules, this effect emerges naturally from end-to-end optimization, providing a principled explanation for improved boundary stability.

In addition, we introduce \textbf{ThyR}, a real thyroid image segmentation dataset consisting of clinically acquired images with expert-annotated boundaries exhibiting realistic ambiguity. The dataset captures challenges such as low contrast, heterogeneous appearance, and uncertain boundaries, offering a valuable benchmark for evaluating uncertainty-aware segmentation methods. The main contributions are summarized as follows:

\begin{itemize}
\item We analyze boundary degradation in encoder--decoder networks from the perspective of sampling-induced ambiguity and intrinsic uncertainty.

\item We propose a noise-injected U-Net framework that perturbs skip connections to regularize cross-scale feature fusion and enhance boundary robustness without additional supervision or architectural complexity.

\item We provide a theoretical interpretation based on local robustness regularization and implicit fuzzy membership modeling.

\item We construct the \textbf{ThyR}\footnote{The ThyR dataset will be released on GitHub upon acceptance.} dataset with expert annotations capturing realistic boundary ambiguity.

\item Extensive experiments demonstrate consistent improvements in segmentation accuracy and boundary quality under challenging conditions.
\end{itemize}

\section{Related Work}

\subsection{U-Net-Based Segmentation and Boundary Limitations}

U-Net \cite{ronneberger2015u} and its variants \cite{zhou2018unet++,valanarasu2022unext,liu2024rolling,ma2024u,li2025u} have achieved remarkable success in image segmentation tasks by leveraging encoder--decoder architectures \cite{kingma2013auto} with skip connections \cite{he2016deep,xie2024usct}. Numerous extensions have been proposed to enhance feature representation, including attention mechanisms \cite{oktay2018attention}, multi-scale fusion \cite{niu2024mind}, and residual learning \cite{he2016deep}. Despite these advances, accurate boundary delineation remains challenging due to information loss during down-sampling and smoothing effects introduced by up-sampling.

Existing boundary-oriented approaches typically rely on explicit supervision, such as auxiliary edge-detection branches or boundary-aware loss functions. Representative examples include boundary-aware segmentation networks, active boundary losses, and edge-supervised segmentation frameworks \cite{qin2021boundary,wang2022active,takikawa2019gated,cheng2021boundary}. While effective, these methods increase model complexity and often require additional annotations or supervision signals. More importantly, they generally treat boundary degradation as a deterministic modeling problem, rather than as a manifestation of uncertainty induced by sampling and image quality limitations.

\subsection{Noise Injection and Robust Learning}

Noise injection has long been recognized as an effective means of improving robustness and generalization in neural networks. Classical techniques such as dropout and stochastic depth introduce stochasticity during training to prevent overfitting and improve optimization stability. More recent studies show that stochastic regularization continues to play an important role in modern deep architectures, particularly in large-scale vision models and data-efficient training regimes \cite{touvron2021training}.

In representation learning, noise and perturbation-based strategies promote invariance by encouraging models to rely on stable and discriminative features. Data augmentation methods such as RandAugment \cite{cubuk2020randaugment} and related policies have demonstrated strong performance improvements across vision tasks. Furthermore, recent work on consistency regularization and perturbation-invariant learning highlights that enforcing stability under input or feature perturbations is crucial for robust dense prediction \cite{french2019semi,mai2024rankmatch}.

Most prior studies focus on Gaussian noise or unbounded perturbations. However, in practical segmentation scenarios, uncertainty is often bounded due to physical sensor limitations and finite pixel resolution. From a robustness perspective, bounded perturbations provide stronger worst-case guarantees and more realistic modeling of local uncertainty \cite{madry2017towards}. Despite these advantages, the role of bounded (e.g., uniform) noise in improving boundary robustness in encoder--decoder segmentation networks remains largely underexplored.

\subsection{Uncertainty and Fuzzy Perspectives}

Fuzzy systems \cite{lu2024fuzzy} have long emphasized the importance of modeling uncertainty and gradual transitions between classes. In image segmentation, boundaries naturally correspond to regions with partial class membership. While fuzzy clustering and rule-based methods explicitly model such uncertainty, their integration with deep encoder--decoder architectures remains limited.

Recent advances in deep learning have increasingly explored uncertainty modeling in segmentation systems. Existing approaches typically estimate uncertainty at the output level through probabilistic predictions, Bayesian inference, or ensemble-based strategies \cite{han2025region,sikha2025uncertainty,karri2025uncertainty}. For example, probabilistic segmentation frameworks and stochastic inference methods aim to quantify prediction confidence and capture ambiguity in complex scenes \cite{chen2024evidence,li2025evidence,li2025evaluation,landgraf2026emuformer}. Related works also explore evidential learning and uncertainty-aware calibration to improve reliability in safety-critical applications \cite{malinin2018predictive}.

Despite these advances, most approaches treat uncertainty as a post-hoc property of model outputs rather than a structural property of intermediate feature representations. Consequently, relatively little attention has been devoted to incorporating uncertainty modeling directly into the feature extraction and feature fusion process of encoder--decoder networks.

This gap motivates the present work, which introduces bounded perturbations into skip connections as an implicit and efficient uncertainty modeling mechanism. From a fuzzy systems perspective, the injected perturbations expand deterministic feature representations into local neighborhoods in latent space, thereby naturally inducing soft class memberships in boundary regions and bridging deep learning with fuzzy representation principles.

\section{Proposed Method}

\begin{figure*}[htbp]
\centering
\includegraphics[width=1\linewidth]{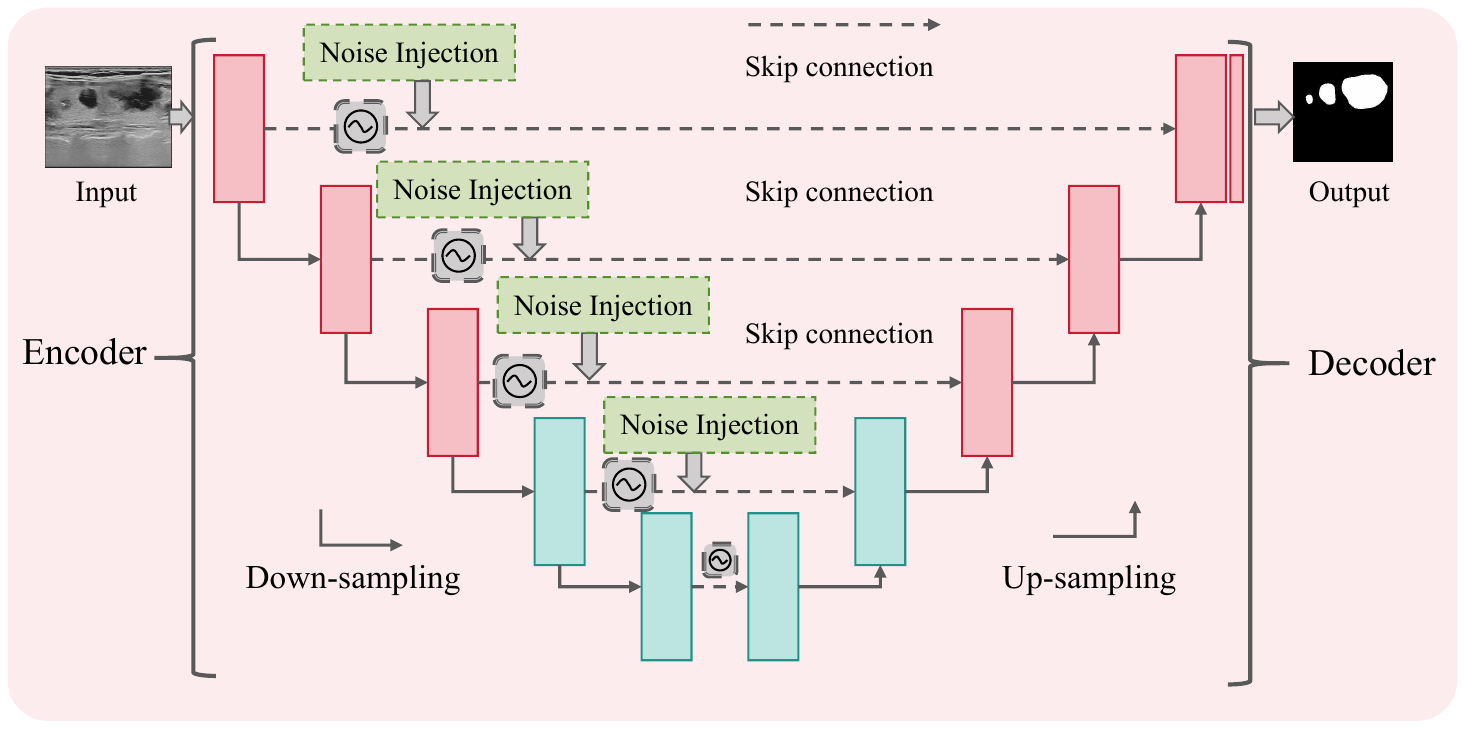}
\caption{Overview of the proposed NoiseUNet framework. A bounded stochastic perturbation is injected into skip-connected feature maps via additive operations to model uncertainty during multi-scale feature fusion. By perturbing cross-scale information flow while preserving the standard concatenation mechanism, the proposed framework enhances boundary robustness without altering the network topology.}
\label{fig_1}
\end{figure*}

\subsection{Baseline U-Net Architecture}

The U-Net architecture consists of an encoder for hierarchical feature extraction and a symmetric decoder for spatial resolution recovery, bridged by skip connections. Given an input image $\mathbf{X}$, the encoder produces a sequence of feature maps $\{\mathbf{F}_l\}_{l=0}^{L}$, where $\mathbf{F}_0 = \mathbf{X}$ and each $\mathbf{F}_l$ is obtained via a convolutional transformation block $\mathcal{E}_l(\cdot)$ followed by a down-sampling operator.

The decoder mirrors the encoder structure and progressively restores spatial resolution using up-sampling operations interleaved with convolutional refinement layers. At each decoding stage, the up-sampled feature map is concatenated with the corresponding encoder feature map via a skip connection:
\begin{equation}
\mathbf{G}_l = \text{Concat}\big( \mathcal{U}_{l+1}(\mathbf{G}_{l+1}), \mathbf{F}_l \big).
\end{equation}

This cross-scale feature fusion enables the decoder to leverage both high-level semantic context and low-level spatial detail for accurate segmentation.

The network outputs a pixel-wise probability map
\begin{equation}
\hat{\mathbf{Y}} = \text{U-Net}_{\boldsymbol{\theta}}(\mathbf{X}),
\end{equation}
where $\boldsymbol{\theta}$ denotes all trainable parameters. The model is trained end-to-end using the pixel-wise cross-entropy loss.

\subsection{Framework Overview}

We build upon the canonical U-Net architecture and introduce a noise-aware modification on skip connections. Rather than modifying the fusion operator itself, we inject bounded stochastic perturbations into skip-connected feature maps before feature fusion, as illustrated in Fig.~\ref{fig_1}.

This design preserves the standard concatenation-based fusion mechanism while introducing uncertainty into cross-scale feature transmission. Since skip connections carry high-resolution spatial features that are critical for boundary reconstruction, perturbing this pathway encourages the network to learn representations that are robust to local inconsistencies and feature ambiguity. The proposed design can be viewed as a minimal yet effective intervention on feature transmission, rather than a modification of network topology.

\subsection{Noise Injection on Skip Connections}

Let $\mathbf{F}_l$ denote the output feature map of the $l$-th encoder block:
\begin{equation}
\mathbf{F}_l = \mathcal{E}_l(\mathbf{F}_{l-1}), \quad \mathbf{F}_0 = \mathbf{X}.
\end{equation}

In the standard U-Net, $\mathbf{F}_l$ is directly forwarded to the decoder via skip connections. In contrast, we introduce bounded stochastic perturbations through an additive operation:
\begin{equation}
\tilde{\mathbf{F}}_l = \mathbf{F}_l + \boldsymbol{\epsilon}_l,
\end{equation}
where $\boldsymbol{\epsilon}_l$ is a stochastic perturbation satisfying
\begin{equation}
\|\boldsymbol{\epsilon}_l\|_\infty \leq \delta.
\end{equation}

The perturbation is applied element-wise across spatial locations and channels. Importantly, no specific distributional assumption is required, making the mechanism distribution-agnostic.

The perturbed feature $\tilde{\mathbf{F}}_l$ is then fused with the decoder feature using the standard concatenation operation:
\begin{equation}
\mathbf{G}_l = \text{Concat}\big( \mathcal{U}_{l+1}(\mathbf{G}_{l+1}), \tilde{\mathbf{F}}_l \big).
\end{equation}

Therefore, the proposed method introduces additive perturbations on skip features while preserving the original U-Net fusion strategy.

\subsection{Effect on Boundary Robustness}

Skip connections play a crucial role in reconstructing object boundaries by transmitting high-resolution spatial details to the decoder. However, these features are often sensitive to noise, low contrast, and misalignment, especially near boundary regions.

The proposed additive perturbation regularizes this process by exposing the decoder to a neighborhood of plausible feature realizations rather than a single deterministic input. In homogeneous regions, small perturbations have minimal impact, whereas in boundary regions characterized by high feature gradients, the perturbations induce noticeable variability. This encourages the network to learn boundary representations that are stable under local uncertainty.

Furthermore, injecting perturbations across multiple skip levels introduces scale-aware robustness. Shallow skip connections address fine-grained edge ambiguity, while deeper ones improve semantic consistency across larger regions. This multi-scale regularization enhances boundary clarity without requiring explicit edge supervision.

\subsection{Fuzzy Interpretation}

From a fuzzy systems perspective, the additive perturbation on skip-connected features induces an implicit fuzzification process. Each deterministic feature vector $\mathbf{f}$ is expanded into a bounded neighborhood:
\begin{equation}
\{\mathbf{f} + \boldsymbol{\epsilon} : \|\boldsymbol{\epsilon}\|_\infty \leq \delta\}.
\end{equation}

This formulation naturally corresponds to soft class membership in boundary regions, where crisp assignments are inherently ambiguous. The fuzzification emerges implicitly through feature perturbation and end-to-end optimization, rather than explicit membership function design.

\subsection{Decoder Architecture and End-to-End Training}

The decoder architecture remains identical to the standard U-Net. No additional parameters, auxiliary branches, or structural modifications are introduced.

Training is performed by minimizing the expected segmentation loss:
\begin{equation}
\mathcal{L} = \mathbb{E}_{\boldsymbol{\epsilon}} \left[
\ell\big( \text{U-Net}_{\boldsymbol{\theta}}(\mathbf{X}; \boldsymbol{\epsilon}), \mathbf{Y} \big)
\right],
\end{equation}
where the expectation is approximated via Monte Carlo sampling.

The bounded perturbation ensures stable optimization and prevents excessive distortion of feature representations. Notably, the proposed method does not require boundary annotations, auxiliary loss functions, or post-processing steps, while effectively improving robustness in boundary-sensitive segmentation tasks.

\section{Theoretical Analysis}

\subsection{Boundary Ambiguity in Cross-Scale Feature Fusion}

In encoder--decoder segmentation networks, object boundaries are primarily reconstructed through the fusion of multi-scale features via skip connections. While encoder features preserve high-resolution spatial details, decoder features encode coarse semantic context. The concatenation of these heterogeneous representations is inherently ill-posed near object boundaries, where spatial alignment is uncertain and semantic transitions are abrupt.

Specifically, encoder features near boundaries often contain high-frequency but noisy responses, whereas decoder features provide smooth but spatially imprecise guidance. Direct fusion of these signals through deterministic skip connections can amplify minor inconsistencies, leading to unstable boundary predictions. This issue becomes more severe under degraded imaging conditions such as blur, noise, and quantization.

Therefore, boundary degradation in U-shaped architectures is not solely caused by down-sampling, but also stems from over-confident cross-scale feature fusion in skip connections.

\subsection{Bounded Additive Perturbation as Skip Fusion Regularization}

Let $\mathbf{F}_l$ denote the encoder feature at level $l$. In the proposed method, a bounded stochastic perturbation is applied via an additive operation:
\begin{equation}
\tilde{\mathbf{F}}_l = \mathbf{F}_l + \boldsymbol{\epsilon}_l,
\end{equation}
where each element of $\boldsymbol{\epsilon}_l$ is independently sampled from a uniform distribution
\begin{equation}
\boldsymbol{\epsilon}_l \sim \mathcal{U}(-\delta, \delta),
\end{equation}
with $\delta = 0.5$ in our implementation, and thus $\|\boldsymbol{\epsilon}_l\|_\infty \leq \delta$.

The perturbed feature $\tilde{\mathbf{F}}_l$ is then concatenated with the decoder feature and processed by subsequent convolutional layers. The training objective becomes
\begin{equation}
\min_{\boldsymbol{\theta}} \; \mathbb{E}_{\boldsymbol{\epsilon}} \left[
\ell\big( \text{U-Net}_{\boldsymbol{\theta}}(\mathbf{X}; \boldsymbol{\epsilon}), \mathbf{Y} \big)
\right].
\end{equation}

This formulation enforces robustness of the decoder with respect to bounded uncertainty in skip-connected features. In effect, optimization is performed over the neighborhood
\begin{equation}
\mathcal{B}_\infty(\mathbf{F}_l, \delta) = \{ \mathbf{F}_l + \boldsymbol{\epsilon} : \|\boldsymbol{\epsilon}\|_\infty \leq \delta \},
\end{equation}
which regularizes the cross-scale fusion process.

Under standard assumptions on convolutional networks, the decoder mapping is locally Lipschitz continuous. Thus, for any two perturbed inputs $\tilde{\mathbf{F}}_l^{(1)}, \tilde{\mathbf{F}}_l^{(2)}$, we have
\begin{equation}
\| \hat{\mathbf{Y}}^{(1)} - \hat{\mathbf{Y}}^{(2)} \|
\leq L \cdot \| \tilde{\mathbf{F}}_l^{(1)} - \tilde{\mathbf{F}}_l^{(2)} \|
\leq 2L\delta,
\end{equation}
where $L$ is the Lipschitz constant. This ensures bounded sensitivity of the decoder to perturbations in skip features.

\subsection{Gradient Smoothing Effect}

Consider the gradient of the expected loss with respect to $\mathbf{F}_l$:
\begin{equation}
\nabla_{\mathbf{F}_l} \mathbb{E}_{\boldsymbol{\epsilon}}[\ell]
= \mathbb{E}_{\boldsymbol{\epsilon}} \left[ \nabla_{\tilde{\mathbf{F}}_l} \ell \right].
\end{equation}

This expectation introduces a smoothing effect on gradient propagation. In boundary regions, where gradients are highly variable due to feature inconsistency across scales, averaging over perturbations suppresses unstable updates and prevents overfitting to spurious high-frequency artifacts.

Let $\mathbf{g}(\boldsymbol{\epsilon}) = \nabla_{\tilde{\mathbf{F}}_l} \ell$. Then the variance satisfies
\begin{equation}
\mathrm{Var}\big( \mathbb{E}_{\boldsymbol{\epsilon}}[\mathbf{g}] \big)
\leq \mathbb{E}_{\boldsymbol{\epsilon}} \left[ \| \mathbf{g}(\boldsymbol{\epsilon}) - \mathbb{E}[\mathbf{g}] \|^2 \right],
\end{equation}
indicating reduced gradient variance when the loss landscape is smooth over $\mathcal{B}_\infty$.

\subsection{Robustness of Skip-Based Fusion}

\textbf{Proposition 1.}
\emph{
Assume the decoder fusion function at level $l$ is locally Lipschitz continuous with constant $L_l$. Then, for additive perturbation $\boldsymbol{\epsilon}_l$ with $\|\boldsymbol{\epsilon}_l\|_\infty \leq \delta$, the variation of the decoder output is bounded by $L_l \delta$.
}

\emph{Proof.}
Let $\Phi_l(\cdot)$ denote the fusion operator (concatenation followed by convolution). Then
\begin{equation}
\| \Phi_l(\mathbf{F}_l + \boldsymbol{\epsilon}_l) - \Phi_l(\mathbf{F}_l) \|
\leq L_l \| \boldsymbol{\epsilon}_l \|
\leq L_l \delta.
\end{equation}
\hfill$\square$

\textbf{Remark 1.}
\emph{
The perturbation is applied additively to skip features prior to concatenation, thereby directly regularizing the cross-scale fusion stage rather than altering the encoder or decoder mappings. This ensures that robustness is enforced precisely where boundary reconstruction occurs.
}

\textbf{Remark 2.}
\emph{
The use of uniform bounded noise provides a worst-case robustness guarantee over $\mathcal{B}_\infty$, avoiding reliance on Gaussian assumptions and ensuring stability under arbitrary bounded perturbations.
}

\subsection{Implicit Fuzzy Membership Induced by Uniform Perturbation}

The additive uniform perturbation transforms each feature vector $\mathbf{f}$ into a local neighborhood:
\begin{equation}
\mathcal{N}_\delta(\mathbf{f}) = \{ \mathbf{f} + \boldsymbol{\epsilon} : \boldsymbol{\epsilon} \sim \mathcal{U}(-\delta,\delta) \}.
\end{equation}

The expected class probability becomes
\begin{equation}
\bar{y}_c(\mathbf{f}) =
\mathbb{E}_{\boldsymbol{\epsilon}} \left[
\sigma_c\big( \text{Decoder}(\mathbf{f} + \boldsymbol{\epsilon}) \big)
\right].
\end{equation}

This expectation can be interpreted as a convolution of the decision function with a uniform kernel, resulting in a softened decision boundary. Near object boundaries, where class evidence is mixed, $\bar{y}_c(\mathbf{f})$ naturally yields intermediate values, corresponding to fuzzy membership degrees.

\textbf{Remark 3.}
\emph{
Uniform perturbation induces a piecewise-linear smoothing effect on the decision function, which closely resembles fuzzy membership functions with bounded support. Unlike Gaussian smoothing, uniform noise preserves sharp transitions while preventing over-confident predictions, making it particularly suitable for modeling boundary uncertainty.
}

\subsection{Discussion}

The analysis shows that additive bounded perturbations on skip connections act as an effective regularizer for cross-scale feature fusion. The method (1) constrains decoder sensitivity via Lipschitz regularization, (2) stabilizes optimization through gradient smoothing, and (3) induces implicit fuzzy memberships through uniform perturbation. These properties explain the improved boundary accuracy and robustness observed in experiments, while preserving the simplicity of the U-Net architecture.

\section{Experiments}
\subsection{Datasets}
\paragraph{BUSI} The dataset \cite{al2020dataset} consists of ultrasound images depicting normal, benign, and malignant breast lesions, each accompanied by corresponding segmentation masks. For our study, we selected 647 images of benign and malignant breast tumors, all resized to 256 × 256 pixels. This well-annotated and diverse collection supports accurate tumor detection and classification, offering valuable insights for both clinical practitioners and researchers.
\paragraph{GlaS} The GlaS (Gland Segmentation) dataset is a prominent benchmark in computational pathology, originally released for the MICCAI 2015 Gland Segmentation Challenge to advance the automated segmentation of gland structures in histology images. Following the common practice adopted by \cite{liu2024rolling}, we selected 165 frames for our experiments and resized them to 512 × 512 pixels.
\paragraph{ThyR} The ThyR dataset was collected over an 12-month period at the First Affiliated Hospital of Shaoyang University and consists of 205 thyroid ultrasound images, including normal, benign and malignant cases. All images were annotated by experienced clinicians and subsequently resized to a uniform resolution of 256 × 256 pixels for consistent analysis.

\subsection{Implementation details}
We implemented NoiseUNet in PyTorch and trained it on an NVIDIA RTX A6000 GPU. For all datasets, we used a batch size of 8 and an initial learning rate of 1e-4. The Adam optimizer was employed together with a cosine annealing learning rate scheduler, with the learning rate decaying to a minimum of 1e-5. The loss function combined binary cross-entropy and Dice loss. Each dataset was randomly partitioned into 80 percent training and 20 percent validation subsets. To ensure robustness, results are reported as averages over three independent runs with different random seeds (2981, 1187, 6142). Standard data augmentations including random rotation and horizontal and vertical flipping were applied during training, which was conducted for 400 epochs. Model performance was evaluated using IoU, F1 Score, GFLOPs, and parameter count.

\subsection{Performance}
\begin{table*}[htbp]
\centering
\caption{Comparison with state-of-the-art segmentation models across three heterogeneous medical scenarios. Results show the average and standard deviation over three random runs.}
\label{tab:comparison}
\tabcolsep=0.5cm
\renewcommand\arraystretch{1.25}
\begin{tabular}{l c c c c c c}
\toprule
\multirow{2}{*}{Methods} & \multicolumn{2}{c}{BUSI} & \multicolumn{2}{c}{GlaS} & \multicolumn{2}{c}{ThyR} \\
\cmidrule(lr){2-3} \cmidrule(lr){4-5} \cmidrule(lr){6-7}
 & IoU$\uparrow$ & F1$\uparrow$ & IoU$\uparrow$ & F1$\uparrow$ & IoU$\uparrow$ & F1$\uparrow$ \\
\midrule
U-Net \cite{ronneberger2015u} & 63.07$\pm$2.19 & 76.35$\pm$2.28 & 88.15$\pm$0.25 & 93.67$\pm$0.14 & 72.80$\pm$0.54 & 83.94$\pm$0.28 \\
U-Net++ \cite{zhou2018unet++} & 62.42$\pm$0.72 & 75.77$\pm$0.94 & 
88.36$\pm$0.41&
93.78$\pm$0.20 &
70.22$\pm$4.73 & 81.75$\pm$3.44 \\
U-NeXt \cite{valanarasu2022unext} & 61.60$\pm$2.26 & 75.34$\pm$2.34 & 88.20$\pm$0.58 & 93.70$\pm$0.31 & 67.73$\pm$2.03 & 80.23$\pm$1.42 \\
Rolling-UNet \cite{liu2024rolling}& 60.67$\pm$3.11 & 74.52$\pm$3.17 & 86.07$\pm$1.17 & 92.44$\pm$0.73 & 67.20$\pm$3.74 & 79.89$\pm$2.93 \\
U-Mamba \cite{ma2024u} & 58.87$\pm$2.58 & 73.01$\pm$2.49 & 88.09$\pm$0.56 & 93.64$\pm$0.30 & 64.27$\pm$7.31 & 77.72$\pm$5.68 \\

Seg. U-KAN \cite{li2025u} & 62.64$\pm$1.50 & 76.13$\pm$1.68 & 87.82$\pm$0.52 & 93.48$\pm$0.29 & 69.68$\pm$1.93 & 81.66$\pm$1.41 \\
\midrule
\textbf{NoiseUNet (Ours)} & \textbf{64.14$\pm$0.72} & \textbf{77.02$\pm$0.76} & \textbf{88.55$\pm$0.25} & \textbf{93.90$\pm$0.14} & \textbf{74.72$\pm$0.60} & \textbf{85.17$\pm$0.62} \\
\bottomrule
\end{tabular}
\end{table*}

\begin{table*}[htbp]
\centering
\caption{Overall comparison with state-of-the-art segmentation models in terms of efficiency and segmentation performance.}
\label{tab:comparison_efficiency}
  \tabcolsep=0.5cm
    \renewcommand\arraystretch{1.25}
\begin{tabular}{l c c c c}
\toprule
\multirow{2}{*}{Methods} & \multicolumn{2}{c}{Average Seg.} & \multicolumn{2}{c}{Efficiency} \\
\cmidrule(lr){2-3} \cmidrule(lr){4-5}
 & IoU$\uparrow$ & F1$\uparrow$ & Gflops & Params (M) \\
\midrule
U-Net \cite{ronneberger2015u} & 74.67$\pm$0.99 & 84.65$\pm$0.90 & 8.98 & 4.75 \\
U-Net++ \cite{zhou2018unet++} & 73.67$\pm$1.95 & 83.77$\pm$1.53 & 1109 & 36.6 \\
U-NeXt \cite{valanarasu2022unext} & 72.51$\pm$1.62 & 83.09$\pm$1.36 & 4.58 & 1.47 \\
Rolling-UNet \cite{liu2024rolling} & 71.31$\pm$2.67 & 82.28$\pm$2.28 & 16.82 & 1.78 \\
U-Mamba \cite{ma2024u} & 70.41$\pm$3.48 & 81.46$\pm$2.82 & 2087 & 86.3 \\
Seg. U-KAN \cite{li2025u}& 73.38$\pm$1.32 & 83.76$\pm$1.13 & 14.02 & 6.35 \\
\midrule
\textbf{NoiseUNet(Ours)} & \textbf{75.80$\pm$0.52} & \textbf{85.36$\pm$0.51} & \textbf{8.98} & \textbf{4.75} \\
\bottomrule
\end{tabular}
\end{table*}

The quantitative results in Table \ref{tab:comparison} demonstrate that the proposed NoiseUNet consistently achieves the best performance across the three heterogeneous medical datasets. On the BUSI dataset, our method improves the baseline U-Net from 63.07\% to 64.14\% IoU and from 76.35\% to 77.02\% F1, showing improved robustness under low-contrast ultrasound conditions. On the GlaS dataset, where most models already achieve high accuracy, NoiseUNet still yields the best results with 88.55\% IoU and 93.90\% F1, indicating stable gains even under strong baselines. The improvement is more evident on the newly introduced ThyR dataset, where our method reaches 74.72\% IoU and 85.17\% F1, outperforming U-Net by 1.92\% IoU and 1.23\% F1. These results suggest that introducing bounded perturbations during feature propagation effectively enhances segmentation robustness in challenging scenarios with ambiguous boundaries.

\begin{figure*}
    \centering
    \includegraphics[width=1\linewidth]{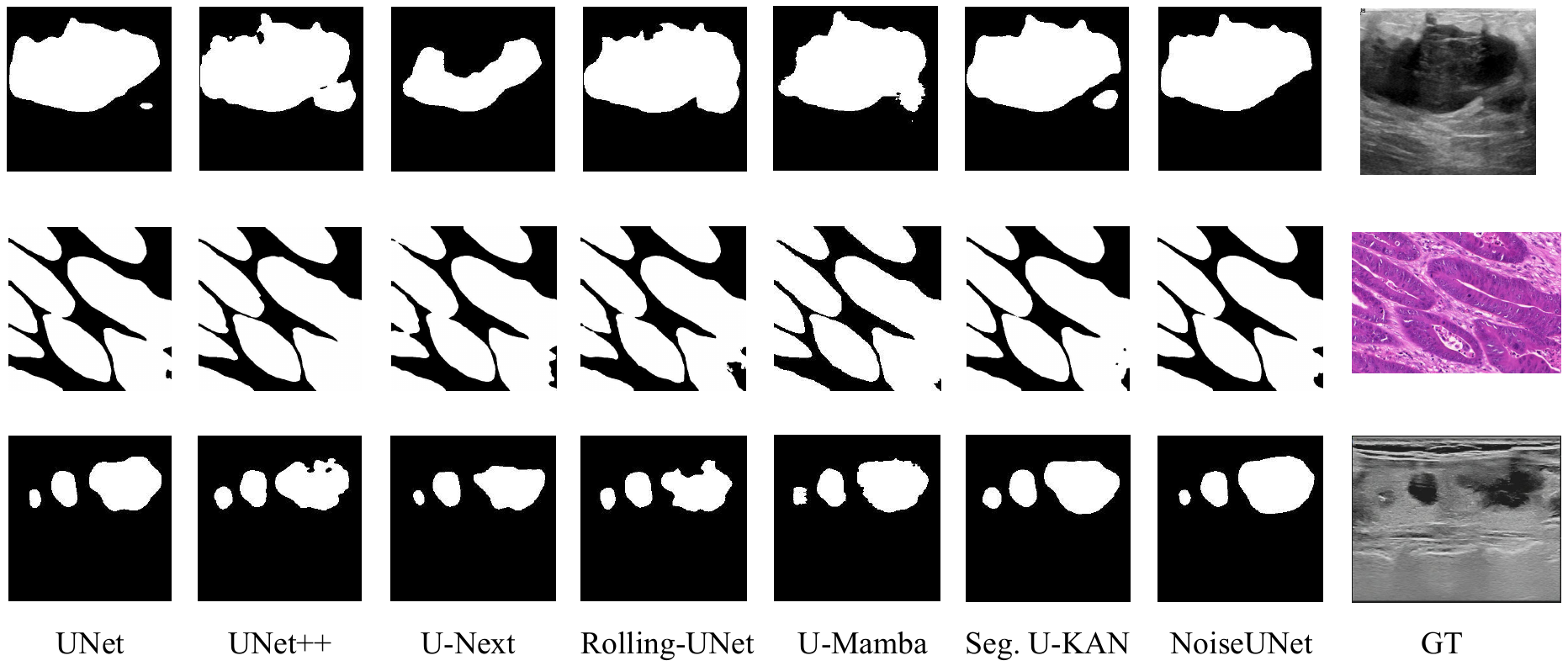}
    \caption{Segmentation results of the proposed NoiseUNet with Skipping Noise on challenging medical images from the BUSI, GlaS, and ThyR datasets.}
    \label{fig:fig3}
\end{figure*}

Table \ref{tab:comparison_efficiency} further shows that the proposed method achieves the highest average segmentation performance (75.80\% IoU and 85.36\% F1) while maintaining the same computational complexity as the standard U-Net (8.98 GFLOPs and 4.75M parameters). This indicates that the performance improvement is obtained without introducing additional architectural overhead. Moreover, the qualitative comparisons in Fig.~\ref{fig:fig3} illustrate that conventional models tend to produce over-smoothed or fragmented boundaries, whereas the proposed method generates clearer and more consistent object contours. This observation is particularly evident on the ThyR dataset, where complex anatomical structures and weak boundary contrast are common, confirming that the proposed perturbation mechanism effectively improves boundary delineation.

\subsection{Ablations}
\subsubsection{encoder-noise vs skip-noise}
\begin{table}[htbp]
\centering
\caption{Performance of Encoder-Noise and Skip-Noise variants.}
\label{tab:niose_xyz}
  \tabcolsep=0.5cm
    \renewcommand\arraystretch{1.25}
\begin{tabular}{c|cc}
\hline
\multirow{2}{*}{NoiseUNet} & \multicolumn{2}{c}{BUSI} \\ \cline{2-3} 
                           & IoU         & F1         \\ \hline
Encoder Noise              &60.35             & 74.36           \\
Skipping Noise   &\textbf{63.33}   &\textbf{76.36}            \\ \hline
NoiseUNet                  & \multicolumn{2}{c}{GlaS} \\ \hline
Encoder Noise              &87.85 &93.45 \\
Skipping Noise & \textbf{88.36}  & \textbf{93.77}  \\ \hline
NoiseUNet                  & \multicolumn{2}{c}{ThyR} \\ \hline
Encoder Noise              &   70.30          & 82.33           \\
Skipping Noise  & \textbf{74.49}  & \textbf{84.97}  \\ \hline
\end{tabular}
\end{table}
The consistent improvement of Skipping Noise over Encoder Noise across all three medical datasets can be understood through the perspective of implicit fuzzification. In Encoder Noise, bounded perturbations are applied only to downsampled encoder features, which have already lost fine boundary details due to pooling or strided convolutions. As a result, the induced uncertainty modeling operates on coarse semantic representations and provides limited regularization for precise contour delineation.

In contrast, Skipping Noise injects bounded noise into skip connections that carry high resolution shallow features preserving spatially accurate boundary information. This ensures that the perturbation acts directly in regions where class membership is inherently ambiguous, such as transition zones between tissues. During training, the decoder fuses these perturbed skip features with upsampled deep features, effectively learning to produce segmentation outputs from a distribution of plausible boundary configurations rather than a single deterministic estimate. This process naturally gives rise to soft, data driven class memberships around object edges, realizing an implicit fuzzification mechanism that is both localized and anatomically meaningful.

Consequently, Skipping Noise achieves more effective uncertainty-aware representation learning at the exact locations where robustness matters most. This explains its higher IoU and F1 scores, especially on challenging datasets like ThyR that exhibit severe boundary ambiguity due to low contrast and imaging artifacts.

\subsubsection{various networks type}
\begin{table}[htbp]
\centering
\caption{Comparison with various base networks.}
\label{tab:network_type}
  \tabcolsep=0.5cm
    \renewcommand\arraystretch{1.25}
\begin{tabular}{cc|cc}
\toprule
\multirow{2}{*}{Networks} & \multirow{2}{*}{Base type} & \multicolumn{2}{c}{ThyR} \\ \cline{3-4} 
& & IoU & F1 \\ \hline
UKAN+Noise & KAN & 69.94 & 82.05\\
UNext+Noise & MLP &68.72 & 81.21 \\
UNet+Noise & CNN & \textbf{74.49} & \textbf{84.97}   \\ \bottomrule
\end{tabular}
\end{table}

We compare three distinct neural architectures that embody fundamentally different inductive biases:

\begin{itemize}
    \item \textbf{Multilayer Perceptron (MLP):}  
    An MLP with $L$ layers computes its output through successive affine transformations followed by element-wise nonlinearities:
    \begin{equation}
        \mathbf{h}^{(0)} = \mathbf{x}, \quad
        \mathbf{h}^{(\ell)} = \sigma\big( \mathbf{W}^{(\ell)} \mathbf{h}^{(\ell-1)} + \mathbf{b}^{(\ell)} \big), \quad \ell = 1, \dots, L,
    \end{equation}
    where $\mathbf{x} \in \mathbb{R}^{d_{\text{in}}}$ is the input, $\mathbf{W}^{(\ell)} \in \mathbb{R}^{d_\ell \times d_{\ell-1}}$ are dense weight matrices, and $\sigma(\cdot)$ is a fixed activation function (e.g., ReLU). The model treats input dimensions as unordered and fully interconnected.

    \item \textbf{Kolmogorov--Arnold Network (KAN):}  
    A KAN replaces linear weights with learnable univariate functions $\phi_{i,j}^{(\ell)}(\cdot)$ acting on scalar inputs. The $\ell$-th layer is defined as:
    \begin{equation}
        h_i^{(0)} = x_i, \quad
        h_i^{(\ell)} = \sum_{j=1}^{n_{\ell-1}} \phi_{j,i}^{(\ell)}\big( h_j^{(\ell-1)} \big), \quad \ell = 1, \dots, L,
    \end{equation}
    where each $\phi_{j,i}^{(\ell)}: \mathbb{R} \to \mathbb{R}$ is a trainable spline-based or parametric function. Unlike MLPs, KANs have no weight matrices; expressivity arises from adaptive basis functions rather than fixed nonlinearities applied to linear combinations.

    \item \textbf{Convolutional Neural Network (CNN):}  
    A CNN processes structured inputs (e.g., images $\mathbf{X} \in \mathbb{R}^{H \times W \times C}$) using local, translation-equivariant filters. The feature map at layer $\ell$ is:
    \begin{equation}
        \mathbf{F}^{(\ell)}_{u,v,k} = \sigma\left( \sum_{c=1}^{C_{\ell-1}} \sum_{p=-P}^{P} \sum_{q=-Q}^{Q} 
        w^{(\ell)}_{p,q,c,k} \cdot \mathbf{F}^{(\ell-1)}_{u+p,\,v+q,\,c} + b^{(\ell)}_k \right),
    \end{equation}
    where $w^{(\ell)}_{p,q,c,k}$ denotes a shared kernel weight applied across all spatial locations $(u,v)$, and $(2P+1) \times (2Q+1)$ is the receptive field size. This enforces locality and parameter efficiency through weight sharing.
\end{itemize}

\noindent These formulations highlight core distinctions: MLPs rely on global dense mixing, KANs on adaptive scalar function composition, and CNNs on local structured operators with spatial invariance.

The results in Table \ref{tab:network_type} show a clear performance hierarchy: CNN-based UNet+Noise significantly outperforms MLP-based UNext+Noise and KAN-based UKAN+Noise on the ThyR dataset in both IoU and F1 score.

This superiority stems from the alignment between convolutional inductive bias and the spatial nature of boundary ambiguity. CNNs exploit local receptive fields and weight sharing, which match the localized uncertainty at object contours. When noise is injected into skip connections, perturbations are propagated in a spatially coherent manner, leading to stable and well-structured boundary representations.

In contrast, MLP-based models lack explicit spatial modeling, causing boundary cues to be diluted under global feature mixing. KAN-based networks, while powerful in functional approximation, do not inherently capture spatial structure or translation equivariance, limiting their ability to leverage noise for boundary refinement. Consequently, UNet+Noise demonstrates that effective implicit fuzzification requires both stochastic perturbation and spatially-aware architectural priors.

\subsubsection{various noise type}
\begin{table}[htbp]
\centering
\caption{Comparison with various noise type.}
\label{tab:noise_type}
  \tabcolsep=0.5cm
    \renewcommand\arraystretch{1.25}
\begin{tabular}{c|cc}
\toprule
\multirow{2}{*}{Noise type} & \multicolumn{2}{c}{ThyR} \\ \cline{2-3} 
                            & IoU         & F1         \\ \hline
Uniform                     & \textbf{74.49}       & \textbf{84.97} \\
Truncated Gaussian          & 74.03       & 84.79 \\
Salt-and-pepper             & 73.69       & 84.66 \\
Poisson-derived variants    & 72.63       & 83.50 \\ \bottomrule
\end{tabular}
\end{table}

Let $\mathbf{F}$ denote the original feature map and $\tilde{\mathbf{F}} = \mathbf{F} + \boldsymbol{\epsilon}$ the perturbed version, where the additive perturbation $\boldsymbol{\epsilon}$ satisfies $\|\boldsymbol{\epsilon}\|_\infty \leq \delta$ for a pre-defined bound $\delta > 0$. We consider the following four bounded stochastic instantiations:

\begin{enumerate}
    \item \textbf{Uniform noise:}
    \begin{equation}
        \boldsymbol{\epsilon}_{ijc} \sim \mathcal{U}(-\delta, \delta), \quad \forall i,j,c.
    \end{equation}

    \item \textbf{Truncated Gaussian noise:}
    \begin{equation}
        \boldsymbol{\epsilon}_{ijc} \sim \mathcal{N}_{[-\delta,\delta]}(0, \sigma^2), \quad \forall i,j,c,
    \end{equation}
    where $\mathcal{N}_{[-\delta,\delta]}(0, \sigma^2)$ denotes a Gaussian distribution with zero mean and variance $\sigma^2$, truncated to the interval $[-\delta, \delta]$.

    \item \textbf{Bounded salt-and-pepper noise:}
    \begin{equation}
        \boldsymbol{\epsilon}_{ijc} =
        \begin{cases}
            +\delta & \text{with probability } p, \\
            -\delta & \text{with probability } p, \\
            0       & \text{with probability } 1 - 2p,
        \end{cases}
        \quad \forall i,j,c,
    \end{equation}
    where $0 < p \leq 0.5$ controls the sparsity of the perturbation.

    \item \textbf{Bounded Poisson-derived noise:}
    \begin{equation}
        \boldsymbol{\epsilon}_{ijc} = \operatorname{clip}\big( k_{ijc} - \lambda,\; -\delta,\; \delta \big), \quad k_{ijc} \sim \operatorname{Poisson}(\lambda),
    \end{equation}
    where $\lambda > 0$ is a fixed rate parameter and $\operatorname{clip}(x, a, b) = \min(\max(x, a), b)$ ensures boundedness.
\end{enumerate}

The results in Table \ref{tab:noise_type} demonstrate the impact of different noise types on segmentation performance using the ThyR dataset. Among the evaluated noise types, uniform noise achieves the highest IoU and F1 scores, followed by truncated Gaussian noise, salt-and-pepper noise, and Poisson-derived variants. This ranking suggests that the distribution and nature of noise significantly influence the model's ability to handle boundary ambiguity. Uniform noise, which introduces perturbations sampled from a continuous interval, effectively models uncertainty in a balanced manner across all spatial locations. This even distribution ensures that each pixel receives equal consideration, promoting robustness against minor variations in intensity or texture, which is crucial for medical image segmentation.

In comparison, truncated Gaussian noise provides perturbations centered around zero but with normal distribution within bounds, leading to slightly less pronounced fuzzification effects than uniform noise. Salt-and-pepper noise, which randomly sets pixels to extreme values or leaves them unchanged, introduces significant local disruptions that may not align well with the smooth transitions typical of anatomical boundaries. This sparse and high-magnitude perturbation can disrupt the learning of fine boundary details, resulting in lower performance. Poisson-derived noise, derived from a fixed-rate process and clipped to ensure boundedness, introduces discrete and potentially non-uniform perturbations that do not naturally fit the continuous structure of medical images. This mismatch between the noise distribution and the data distribution leads to reduced effectiveness in modeling boundary ambiguity, as evidenced by the lowest performance among the tested noise types. Overall, uniform noise stands out as the optimal choice for enhancing segmentation accuracy and robustness in medical image analysis tasks.

\section{Conclusion}

This paper addressed boundary degradation in encoder--decoder segmentation networks from the perspective of cross-scale feature fusion uncertainty. We proposed a simple yet effective noise-injected U-Net framework that introduces bounded additive perturbations on skip-connected features prior to concatenation, thereby regularizing feature fusion without altering network topology. Theoretically, the method enforces local robustness via Lipschitz-constrained mappings and gradient smoothing, while implicitly inducing fuzzy memberships in boundary regions through bounded uniform perturbations. Extensive experiments demonstrate consistent improvements in boundary delineation and overall segmentation performance without additional supervision or parameters, highlighting the effectiveness of modeling uncertainty directly at the feature level. Future work will explore adaptive and spatially-aware perturbation strategies for broader dense prediction tasks.

\section*{Acknowledgments}
This paper was supported by School of Information Science and Engineering, Provincial Key Laboratory of Informational Service for Rural Area of Southwestern Hunan, Shaoyang University, Shaoyang 422000, China.


\bibliographystyle{IEEEtran}
\bibliography{mybibfile}

@inproceedings{ronneberger2015u,
  title={U-net: Convolutional networks for biomedical image segmentation},
  author={Ronneberger, Olaf and Fischer, Philipp and Brox, Thomas},
  booktitle={International Conference on Medical image computing and computer-assisted intervention},
  pages={234--241},
  year={2015},
  organization={Springer}
}

@inproceedings{zhou2018unet++,
  title={Unet++: A nested u-net architecture for medical image segmentation},
  author={Zhou, Zongwei and Rahman Siddiquee, Md Mahfuzur and Tajbakhsh, Nima and Liang, Jianming},
  booktitle={International workshop on deep learning in medical image analysis},
  pages={3--11},
  year={2018},
  organization={Springer}
}

@inproceedings{valanarasu2022unext,
  title={Unext: Mlp-based rapid medical image segmentation network},
  author={Valanarasu, Jeya Maria Jose and Patel, Vishal M},
  booktitle={International conference on medical image computing and computer-assisted intervention},
  pages={23--33},
  year={2022},
  organization={Springer}
}

@inproceedings{liu2024rolling,
  title={Rolling-unet: Revitalizing mlp’s ability to efficiently extract long-distance dependencies for medical image segmentation},
  author={Liu, Yutong and Zhu, Haijiang and Liu, Mengting and Yu, Huaiyuan and Chen, Zihan and Gao, Jie},
  booktitle={Proceedings of the AAAI conference on artificial intelligence},
  volume={38},
  number={4},
  pages={3819--3827},
  year={2024}
}

@article{ma2024u,
  title={U-mamba: Enhancing long-range dependency for biomedical image segmentation},
  author={Ma, Jun and Li, Feifei and Wang, Bo},
  journal={arXiv preprint arXiv:2401.04722},
  year={2024}
}

@inproceedings{li2025u,
  title={U-kan makes strong backbone for medical image segmentation and generation},
  author={Li, Chenxin and Liu, Xinyu and Li, Wuyang and Wang, Cheng and Liu, Hengyu and Liu, Yifan and Chen, Zhen and Yuan, Yixuan},
  booktitle={Proceedings of the AAAI Conference on Artificial Intelligence},
  volume={39},
  number={5},
  pages={4652--4660},
  year={2025}
}

@article{kingma2013auto,
  title={Auto-encoding variational bayes},
  author={Kingma, Diederik P and Welling, Max},
  journal={arXiv preprint arXiv:1312.6114},
  year={2013}
}

@article{xie2024usct,
  title={USCT-UNet: Rethinking the semantic gap in U-Net network from U-shaped skip connections with multichannel fusion transformer},
  author={Xie, Xiaoshan and Yang, Min},
  journal={IEEE Transactions on Neural Systems and Rehabilitation Engineering},
  volume={32},
  pages={3782--3793},
  year={2024},
  publisher={IEEE}
}

@inproceedings{he2016deep,
  title={Deep residual learning for image recognition},
  author={He, Kaiming and Zhang, Xiangyu and Ren, Shaoqing and Sun, Jian},
  booktitle={Proceedings of the IEEE conference on computer vision and pattern recognition},
  pages={770--778},
  year={2016}
}

@article{oktay2018attention,
  title={Attention u-net: Learning where to look for the pancreas},
  author={Oktay, Ozan and Schlemper, Jo and Folgoc, Loic Le and Lee, Matthew and Heinrich, Mattias and Misawa, Kazunari and Mori, Kensaku and McDonagh, Steven and Hammerla, Nils Y and Kainz, Bernhard and others},
  journal={arXiv preprint arXiv:1804.03999},
  year={2018}
}

@article{niu2024mind,
  title={Mind the gap: Learning modality-agnostic representations with a cross-modality unet},
  author={Niu, Xin and Li, Enyi and Liu, Jinchao and Wang, Yan and Osadchy, Margarita and Fang, Yongchun},
  journal={IEEE Transactions on Image Processing},
  volume={33},
  pages={655--670},
  year={2024},
  publisher={IEEE}
}

@article{qin2021boundary,
  title={Boundary-aware segmentation network for mobile and web applications},
  author={Qin, Xuebin and Fan, Deng-Ping and Huang, Chenyang and Diagne, Cyril and Zhang, Zichen and Sant'Anna, Adri{\`a} Cabeza and Suarez, Albert and Jagersand, Martin and Shao, Ling},
  journal={arXiv preprint arXiv:2101.04704},
  year={2021}
}

@inproceedings{wang2022active,
  title={Active boundary loss for semantic segmentation},
  author={Wang, Chi and Zhang, Yunke and Cui, Miaomiao and Ren, Peiran and Yang, Yin and Xie, Xuansong and Hua, Xian-Sheng and Bao, Hujun and Xu, Weiwei},
  booktitle={Proceedings of the AAAI conference on artificial intelligence},
  volume={36},
  number={2},
  pages={2397--2405},
  year={2022}
}

@inproceedings{cheng2021boundary,
  title={Boundary IoU: Improving object-centric image segmentation evaluation},
  author={Cheng, Bowen and Girshick, Ross and Doll{\'a}r, Piotr and Berg, Alexander C and Kirillov, Alexander},
  booktitle={Proceedings of the IEEE/CVF conference on computer vision and pattern recognition},
  pages={15334--15342},
  year={2021}
}

@inproceedings{takikawa2019gated,
  title={Gated-scnn: Gated shape cnns for semantic segmentation},
  author={Takikawa, Towaki and Acuna, David and Jampani, Varun and Fidler, Sanja},
  booktitle={Proceedings of the IEEE/CVF international conference on computer vision},
  pages={5229--5238},
  year={2019}
}

@article{han2025region,
  title={Region uncertainty estimation for medical image segmentation with noisy labels},
  author={Han, Kai and Wang, Shuhui and Chen, Jun and Qian, Chengxuan and Lyu, Chongwen and Ma, Siqi and Qiu, Chengjian and Sheng, Victor S and Huang, Qingming and Liu, Zhe},
  journal={IEEE Transactions on Medical Imaging},
  year={2025},
  publisher={IEEE}
}

@article{sikha2025uncertainty,
  title={Uncertainty-aware segmentation quality prediction via deep learning Bayesian modeling: comprehensive evaluation and interpretation on skin cancer and liver segmentation},
  author={Sikha, OK and Riera-Marin, Meritxell and Galdran, Adrian and Lopez, Javier Garcia and Rodriguez-Comas, Julia and Piella, Gemma and Ballester, Miguel A Gonzalez},
  journal={Computerized Medical Imaging and Graphics},
  volume={123},
  pages={102547},
  year={2025},
  publisher={Elsevier}
}

@inproceedings{karri2025uncertainty,
  title={Uncertainty-guided cross attention ensemble mean teacher for semi-supervised medical image segmentation},
  author={Karri, Meghana and Arya, Amit Soni and Biswas, Koushik and Gennaro, Nicolo and Cicek, Vedat and Durak, Gorkem and Velichko, Yury S and Bagci, Ulas},
  booktitle={Proceedings of the Winter Conference on Applications of Computer Vision},
  pages={7039--7048},
  year={2025}
}

@inproceedings{li2025evidence,
  title={An Evidence-Based Tri-Branch Cross-Pseudo Supervision Method for Semi-Supervised Medical Image Segmentation},
  author={Li, Dongyue and Luo, Aocheng and Wang, Shaoan and Hu, Yaoqing and Pan, Jie and Wang, Yifei and Yu, Junzhi},
  booktitle={2025 IEEE/RSJ International Conference on Intelligent Robots and Systems (IROS)},
  pages={7733--7738},
  year={2025},
  organization={IEEE}
}

@article{li2025evaluation,
  title={Evaluation of uncertainty estimation methods in medical image segmentation: Exploring the usage of uncertainty in clinical deployment},
  author={Li, Shiman and Yuan, Mingzhi and Dai, Xiaokun and Zhang, Chenxi},
  journal={Computerized Medical Imaging and Graphics},
  volume={124},
  pages={102574},
  year={2025},
  publisher={Elsevier}
}

@article{landgraf2026emuformer,
  title={EMUFormer: Efficient Multi-task Uncertainties for Reliable Joint Semantic Segmentation and Monocular Depth Estimation},
  author={Landgraf, Steven and Hillemann, Markus and Kapler, Theodor and Ulrich, Markus},
  journal={International Journal of Computer Vision},
  volume={134},
  number={4},
  pages={142},
  year={2026},
  publisher={Springer}
}

@article{chen2024evidence,
  title={Evidence-based uncertainty-aware semi-supervised medical image segmentation},
  author={Chen, Yingyu and Yang, Ziyuan and Shen, Chenyu and Wang, Zhiwen and Zhang, Zhongzhou and Qin, Yang and Wei, Xin and Lu, Jingfeng and Liu, Yan and Zhang, Yi},
  journal={Computers in Biology and Medicine},
  volume={170},
  pages={108004},
  year={2024},
  publisher={Elsevier}
}

@article{lu2024fuzzy,
  title={Fuzzy machine learning: A comprehensive framework and systematic review},
  author={Lu, Jie and Ma, Guangzhi and Zhang, Guangquan},
  journal={IEEE Transactions on Fuzzy Systems},
  volume={32},
  number={7},
  pages={3861--3878},
  year={2024},
  publisher={IEEE}
}

@inproceedings{touvron2021training,
  title={Training data-efficient image transformers \& distillation through attention},
  author={Touvron, Hugo and Cord, Matthieu and Douze, Matthijs and Massa, Francisco and Sablayrolles, Alexandre and J{\'e}gou, Herv{\'e}},
  booktitle={International conference on machine learning},
  pages={10347--10357},
  year={2021},
  organization={PMLR}
}

@inproceedings{cubuk2020randaugment,
  title={Randaugment: Practical automated data augmentation with a reduced search space},
  author={Cubuk, Ekin D and Zoph, Barret and Shlens, Jonathon and Le, Quoc V},
  booktitle={Proceedings of the IEEE/CVF conference on computer vision and pattern recognition workshops},
  pages={702--703},
  year={2020}
}

@article{french2019semi,
  title={Semi-supervised semantic segmentation needs strong, varied perturbations},
  author={French, Geoff and Laine, Samuli and Aila, Timo and Mackiewicz, Michal and Finlayson, Graham},
  journal={arXiv preprint arXiv:1906.01916},
  year={2019}
}

@inproceedings{mai2024rankmatch,
  title={Rankmatch: Exploring the better consistency regularization for semi-supervised semantic segmentation},
  author={Mai, Huayu and Sun, Rui and Zhang, Tianzhu and Wu, Feng},
  booktitle={Proceedings of the IEEE/CVF conference on computer vision and pattern recognition},
  pages={3391--3401},
  year={2024}
}

@article{madry2017towards,
  title={Towards deep learning models resistant to adversarial attacks},
  author={Madry, Aleksander and Makelov, Aleksandar and Schmidt, Ludwig and Tsipras, Dimitris and Vladu, Adrian},
  journal={arXiv preprint arXiv:1706.06083},
  year={2017}
}

@article{malinin2018predictive,
  title={Predictive uncertainty estimation via prior networks},
  author={Malinin, Andrey and Gales, Mark},
  journal={Advances in neural information processing systems},
  volume={31},
  year={2018}
}

@article{liao2024double,
  title={Double integral-enhanced Zeroing neural network with linear noise rejection for time-varying matrix inverse},
  author={Liao, Bolin and Han, Luyang and Cao, Xinwei and Li, Shuai and Li, Jianfeng},
  journal={CAAI Transactions on Intelligence Technology},
  volume={9},
  number={1},
  pages={197--210},
  year={2024},
  publisher={Wiley Online Library}
}

@article{wang2025prescribed,
  title={Prescribed-time convergence noise-tolerant zeroing neural network for multi-robot position management and coordination},
  author={Wang, Tinglei and Wang, Yufei and Hua, Cheng and Cao, Xinwei and Liao, Bolin and Li, Shuai},
  journal={Engineering Applications of Artificial Intelligence},
  volume={162},
  pages={112072},
  year={2025},
  publisher={Elsevier}
}

@article{li2024review,
  title={A review of remote sensing image segmentation by deep learning methods},
  author={Li, Jiangyun and Cai, Yuanxiu and Li, Qing and Kou, Mingyin and Zhang, Tianxiang},
  journal={International Journal of Digital Earth},
  volume={17},
  number={1},
  pages={2328827},
  year={2024},
  publisher={Taylor \& Francis}
}

@article{wu2025farmseg_vlm,
  title={FarmSeg\_VLM: A farmland remote sensing image segmentation method considering vision-language alignment},
  author={Wu, Haiyang and Mu, Weiliang and Zhong, Dandan and Du, Zhuofei and Li, Haifeng and Tao, Chao},
  journal={ISPRS Journal of Photogrammetry and Remote Sensing},
  volume={225},
  pages={423--439},
  year={2025},
  publisher={Elsevier}
}

@article{zhang2025small,
  title={Small object few-shot segmentation for vision-based industrial inspection},
  author={Zhang, Zilong and Niu, Chang and Zhao, Zhibin and Zhang, Xingwu and Chen, Xuefeng},
  journal={IEEE Transactions on Industrial Informatics},
  year={2025},
  publisher={IEEE}
}

@article{wang2025overview,
  title={An overview of industrial image segmentation using deep learning models},
  author={Wang, Guina and Li, Zhen and Weng, Guirong and Chen, Yiyang},
  journal={Intelligence \& Robotics},
  volume={5},
  number={1},
  pages={143--180},
  year={2025},
  publisher={OAE Publishing Inc.}
}

@article{al2020dataset,
  title={Dataset of breast ultrasound images},
  author={Al-Dhabyani, Walid and Gomaa, Mohammed and Khaled, Hussien and Fahmy, Aly},
  journal={Data in brief},
  volume={28},
  pages={104863},
  year={2020},
  publisher={Elsevier}
}

\vfill

\end{document}